\documentclass{poleval}

\usepackage{layout}

\begin{document}

\chapter{Approaching English-Polish Machine Translation Quality Assessment with Neural-based Methods}

\fancyhead[LO]{\textnormal{\small{Approaching English-Polish Machine Translation Quality Assessment with Neural-based Methods}}}

\chapterauthor[Artur Nowakowski]{Artur Nowakowski \textnormal{(Faculty of Mathematics and Computer Science, Adam Mickiewicz University, Pozna\'n)}}


\begin{abstract}
This paper presents our contribution to the PolEval 2021 Task 2: \textit{Evaluation of translation quality assessment metrics}. 
We describe experiments with pre-trained language models and state-of-the-art frameworks for translation quality assessment in both \textit{nonblind} and \textit{blind} versions of the task. 
Our solutions ranked second in the \textit{nonblind} version and third in the \textit{blind} version.

\end{abstract}


\begin{keywords}
machine translation quality estimation, machine translation evaluation, pre-trained language models, natural language processing 
\end{keywords}


\section{Introduction}
Machine translation quality evaluation is the task of assessing translation quality based on a reference translation. 
In the past, traditional machine translation evaluation metrics such as \textsc{bleu} \citep{papineni-etal-2002-bleu}, \textsc{meteor} \citep{banerjee-lavie-2005-meteor}, or \textsc{chrF} \citep{popovic-2015-chrf} relied on lexical-level features between the machine translation hypothesis and the reference translation. 
They remain popular to this day due to their computational speed and the fact that they can be applied to any translation direction.

The rise of Neural Machine Translation (NMT) in recent years has shown that high-quality NMT systems are often mistreated by lexical-level evaluation metrics, as such systems can generate correct translation that is lexically distant from a reference translation. 

Recent advances in the field of neural language modeling \citep{devlin-etal-2019-bert, conneau2020unsupervised} led to the creation of BERT cosine similarity-based metrics, such as \textsc{BERTScore} \citep{bert-score}, as well as metrics trained on human judgments, such as \textsc{Comet} \citep{rei-etal-2020-comet} and \textsc{Bleurt} \citep{sellam-etal-2020-bleurt}.
Human judgments include manually assigned quality scores, such as \textit{Direct Assessment} (DA) \citep{graham-etal-2013-continuous}, but may also be derived from post-edited translation to calculate post-editing effort in the form of \textit{Human-mediated Translation Edit Rate} (HTER) \citep{snover-etal-2006-study}.

Machine translation quality estimation (QE) is a different task than evaluation, as the goal is to predict machine translation quality without access to a reference translation. 
Research on QE in recent years has shown that it is possible to achieve high levels of correlation with human judgments based only on a source segment and a machine translation hypothesis \citep{specia-etal-2020-findings-wmt}.
Existing state-of-the-art frameworks for QE include \textsc{Comet} \citep{rei-etal-2020-comet}, which allows QE models to be trained in a reference-free mode and \textsc{TransQuest} \citep{transquest:2020a}, which proposes two new architectures for QE: \textsc{MonoTransQuest} and \textsc{SiameseTransQuest}.


\section{Task description}
The goal of Task 2 is to investigate metrics for automatic evaluation of machine translation in the English-Polish translation direction.

The organizers prepared distinct datasets for \textit{nonblind} and \textit{blind} versions of the task. The \textit{nonblind} dataset consists of the following data: source segment, machine translation hypothesis, reference translation, and quality score. 
The \textit{blind} dataset consists only of machine translation hypothesis and its quality score.
The segment quality scores were created by averaging the scores assigned by six human annotators.
Unlike most of the current human judgment-based QE tasks, where scores are assigned on a continuous scale \citep{graham-etal-2013-continuous}, the task utilizes a standard Likert scale allowing ratings from 1 to 5.  
The evaluation metric used in both versions of the task is Pearson's \textit{r} correlation score.

The datasets were split into a development set ("dev-0") and two test sets ("test-A" and "test-B"). 
The first of the test sets ("test-A") was the main test set during the initial testing phase of the competition and was converted to the development set with the release of the final test set ("test-B").

Table \ref{dataset_statistics} presents statistics of the provided datasets: the number of segments, the average number of source tokens, the average number of MT hypothesis tokens, the minimum segment quality score, and the average segment quality score.

\begin{table}[htbp]
\centering
\caption{Statistics of datasets provided by organizers.}
\label{dataset_statistics}
\begin{tabular}{lrrrrrr}
\hline
\multicolumn{1}{c}{}                 & \multicolumn{3}{c}{\textbf{Nonblind}} & \multicolumn{3}{c}{\textbf{Blind}} \\
\multicolumn{1}{c}{}                 & \textbf{Dev-0}   & \textbf{Test-A}   & \textbf{Test-B}  & \textbf{Dev-0}  & \textbf{Test-A}  & \textbf{Test-B} \\ \hline
Segments                   & 485         &    500      &    1000     & 485       &  500       &    1000    \\
Avg. tokens (source) & 18.22 &   17.36       &   17.73      &  -       &    -     &     -   \\
Avg. tokens (MT hypothesis) & 16.23  &  15.49       &  15.78    & 17.55       & 16.49        & 16.57        \\
Min. score                       & 3.0        & 2.58         & 2.92       & 3.08       & 2.67     & 2.0        \\
Avg. score                       & 4.30      &  4.37         & 4.38        & 4.33        &  4.31       & 4.40        \\ \hline
\end{tabular}
\end{table}

\section{Solutions}
\subsection{\textit{Nonblind} task version solution}
Our final solution to the \textit{nonblind} version of the task is based on \textsc{Comet}. 
We used the "test-A" dataset as the training data and the "dev-0" dataset as the development data. 

\textsc{Comet} uses pre-trained language model as the encoder for the source segment, the machine translation hypothesis, and the reference translation, which are independently encoded.
Therefore, we decided to use HerBERT\textsubscript{LARGE} \citep{mroczkowski-etal-2021-herbert} as the pre-trained encoder model. 
We also experimented with XLM-RoBERTa \citep{conneau2020unsupervised} (XLM-R) as the pre-trained encoder model, but the results were subpar.
It is because HerBERT\textsubscript{LARGE} model was trained specifically for the Polish language and initialized with XLM-RoBERTa weights.

We applied gradual unfreezing and discriminative learning rates \citep{howard-ruder-2018-universal}, meaning that we kept the encoder model frozen for 8 epochs while the feed-forward regressor was optimized with the learning rate of \(3\mathrm{e}{-5}\). 
After 8 epochs, the entire model is fine-tuned but the learning rate is reduced to \(1\mathrm{e}{-5}\) to avoid catastrophic forgetting. All hyperparameters used for training \textsc{Comet} models are presented in Table \ref{comet_hyperparameters}.

We experimented with other state-of-the-art methods for machine translation evaluation as well. 
We used \textsc{BERTScore} with contextual embeddings from the HerBERT\textsubscript{LARGE} model and found that it generates promising results given that it is based on cosine similarity and is not fine-tuned on the task data in any way. 

Out of the trained metrics, we also experimented with \textsc{Bleurt} and \textsc{TransQuest} with \textsc{MonoTransQuest} architecture. 
The \textsc{Bleurt} model was fine-tuned on the open-source \textit{bleurt-base-128} model\footnote{https://github.com/google-research/bleurt/blob/master/checkpoints.md} with default hyperparameters. 
The \textsc{TransQuest} model was fine-tuned on the open-source English-to-Any model pre-trained on DA\footnote{https://tharindu.co.uk/TransQuest/models/sentence\_level\_pretrained} with default hyperparameters. 
\textsc{TransQuest} is trained only on the source segment and the machine translation hypothesis and does not take into account the reference translation. 
The final results of all methods used in the \textit{nonblind} version of the task are presented in Table \ref{nonblind_results}.

\begin{table}[htbp]
\centering
\caption{Results of the \textit{nonblind} version of the task on the "test-B" dataset.}
\label{nonblind_results}
\begin{tabular}{lc}
\hline
Method          & Pearson's \textit{r} \\ \hline
\textsc{Comet} (HerBERT) & \textbf{57.28}            \\
\textsc{Comet} (XLM-R)   & 53.84            \\
\textsc{Bleurt}          & 57.25           \\
\textsc{TransQuest}      & 55.70           \\
\textsc{BERTScore}       & 48.74  \\ \hline
\end{tabular}
\end{table}

\subsection{\textit{Blind} task version solution}
Our final solution to the \textit{blind} version of the task is based on \textsc{Comet} as well.

The provided dataset contains only machine translation hypotheses in this scenario.
Therefore, we decided to create synthetic source segments by back-translating the provided machine translation hypotheses into English by using the open-source OPUS-MT \citep{TiedemannThottingal:EAMT2020} NMT model\footnote{https://huggingface.co/Helsinki-NLP/opus-mt-pl-en}, which is based on the Marian \citep{mariannmt} framework.

We combined all the data from the \textit{nonblind} dataset with the back-translated data from the \textit{blind} dataset. 
Then, we randomly selected 100 segment pairs as the development set.

The model training procedure is the same as in the \textit{nonblind} solution. 
The only difference is that the \textsc{Comet} model was trained in the reference-free mode in this scenario.
Hyperparameters used for the \textit{blind} model training are presented in Table \ref{comet_hyperparameters}.

In this version of the task, we also conducted experiments using \textsc{TransQuest}. \textsc{TransQuest} model architecture, hyperparameters, and used pre-trained model were the same as in the solution to the \textit{nonblind} version of the task. 
The final results of all methods used in the \textit{blind} version of the task are presented in Table \ref{blind_results}.

\begin{table}[htbp]
\centering
\caption{Results of the \textit{blind} version of the task on the "test-B" dataset.}
\label{blind_results}
\begin{tabular}{lc}
\hline
Method          & Pearson's \textit{r} \\ \hline
\textsc{Comet} (HerBERT) & \textbf{47.93}            \\
\textsc{Comet} (XLM-R)   & 43.52            \\
\textsc{TransQuest}      & 41.71          \\ \hline
\end{tabular}
\end{table}

\begin{table}[htbp]
\centering
\caption{Hyperparameters used for training \textsc{Comet} models.}
\label{comet_hyperparameters}
\begin{tabular}{lcc}
\hline
Hyperparameter  & Nonblind model & Blind model \\ \hline
Pre-trained encoder model &  HerBERT\textsubscript{LARGE} & HerBERT\textsubscript{LARGE}  \\
Optimizer   &   Adam (default parameters)  & AdamW (default parameters)             \\
Learning rate  & \(3\mathrm{e}{-5}\) and \(1\mathrm{e}{-5}\)  & \(3.1\mathrm{e}{-5}\) and \(1\mathrm{e}{-5}\)  \\
Layer-wise decay  & -  & 0.95  \\ 
Num. of frozen epochs    &  8           &  0.3           \\
Batch size  & 4     &  2 \\
Accumulated gradient batches & 2     &  4 \\
Loss function & MSE                & MSE  \\
Dropout  & 0.15               &  0.15 \\
Feed-forward hidden units  & 4096, 2048               & 2048, 1024  \\
Feed-forward activation function          & Tanh    & Tanh \\ \hline
\end{tabular}
\end{table}

\section{Conclusions}
We presented our contribution to the PolEval 2021 Task 2: \textit{Evaluation of translation quality assessment metrics}. 

The experiments consisted in comparing state-of-the-art methods for translation quality assessment in the English-Polish translation direction.
The final solutions are based on the \textsc{Comet} framework.
The solutions achieved second and third place in the \textit{nonblind} and \textit{blind} versions of the task, respectively.
In the \textit{blind} version of the task, we presented a procedure for creating a synthetic source segment input by back-translating machine translation hypothesis.
All of the described methods are also worth further investigation in future experiments, as they generate competitive results.

The code and models used for creating the solutions are open-source and available on GitHub\footnote{https://github.com/arturnn/poleval2021-qe}. 


\bibliography{poleval}
\bibliographystyle{poleval}

\end{document}